\def\year{2022}\relax
\documentclass[letterpaper]{article} 
\usepackage{aaai22}  
\usepackage{times}  
\usepackage{helvet}  
\usepackage{courier}  
\usepackage[hyphens]{url}  
\usepackage{graphicx} 
\usepackage{natbib}  
\usepackage{caption} 
\DeclareCaptionStyle{ruled}{labelfont=normalfont,labelsep=colon,strut=off} 
\usepackage{algorithm}
\usepackage{algorithmic}
\usepackage{bibentry}
\usepackage{algorithm}
\usepackage{algorithmic}

\usepackage{amsmath}
\usepackage{bm}

\usepackage{microtype}

\usepackage{mathrsfs}

\usepackage{graphicx}
\usepackage{amsmath}
\usepackage{multirow}
\usepackage{mathtools}
\usepackage{placeins}
\usepackage{float}

\usepackage{booktabs}
\usepackage{ragged2e}

\usepackage{graphics}
\usepackage{epsfig}
\usepackage{amssymb}
\usepackage{threeparttable}
\usepackage{stmaryrd}
\usepackage{subcaption}
\usepackage[export]{adjustbox}

\usepackage{bm}
\usepackage{subcaption}
\usepackage{rotating}
\usepackage{stackengine}

\usepackage{adjustbox}

\usepackage{newfloat}
\usepackage{listings}
\floatstyle{ruled}
\newfloat{listing}{tb}{lst}{}
\floatname{listing}{Listing}
\title{Learning Ambiguity from Crowd Sequential Annotations}
\author{
    Xiaolei Lu \textsuperscript{\rm 1}
}
\affiliations{
    \textsuperscript{\rm 1} City University of Hong Kong\\

}

\begin{document}


\maketitle

\begin{abstract}
Most crowdsourcing learning methods treat disagreement between annotators as noisy labelings while inter-disagreement among experts is often a good indicator for the ambiguity and uncertainty that is inherent in natural language. In this paper, we propose a framework called Learning Ambiguity from Crowd Sequential Annotations (LA-SCA) to explore the inter-disagreement between reliable annotators and effectively preserve confusing label information. First, a hierarchical Bayesian model is developed to infer ground-truth from crowds and group the annotators with similar reliability together. By modeling the relationship between the size of group the annotator involved in, the annotator’s reliability and element’s unambiguity in each sequence, inter-disagreement between reliable annotators on ambiguous elements is computed to obtain label confusing information that is incorporated to cost-sensitive sequence labeling. Experimental results on POS tagging and NER tasks show that our proposed framework achieves competitive performance in inferring ground-truth from crowds and predicting unknown sequences, and interpreting hierarchical clustering results helps discover labeling patterns of annotators with similar reliability.

\end{abstract}

\section{Introduction}

Sequence labeling, which refers to assign sequences of labels to observed sequential data, is widely used in Natural Language Processing (NLP) tasks including Part-of-Speech (POS) tagging, Chunking and Named Entity Recognition (NER). Many downstream NLP applications (e.g. relation extraction and machine translation ) can benefit from sequential label assignments of these fundamental NLP tasks.

Traditional sequence labeling models like Hidden Markov Models (HMMs) and Conditional Random Fields (CRFs) require handcrafted features which need to be carefully designed to obtain good results on a specific dataset. Over the past decade, deep sequence models have resulted in improving the performance of sequence labeling. For example, Bi-LSTM-CRF \cite{re1} and Transformer\cite{re2}. However, these sequence labeling models require a large amount of training data with exact annotations, which is costly and laborious to produce. 

In recent years, well-developed commercial crowdsourcing platforms (e.g. Amazon Mechanical Turk and CrowdFlower \cite{re3}) have flourished as effective tools to obtain large labeled datasets. Crowdsourcing utilizes contribution of the group's intelligence, but the quality of crowd labels still cannot be guaranteed as the expertise level of annotators varies. Therefore the major focus of learning from crowds is on estimating the reliability of annotators and building prediction models based on the estimated ground-truth labels. For example, Snow et al. \cite{re4} used bias correction to combine non-expert annotation. Raykar et al. \cite{re5} proposed to jointly estimate the coefficients of a logistic regression classifier and the annotators’ expertise.

Many effective models like HMM-Crowd \cite{re6} and Sembler \cite{re7} extend crowdsourcing to sequence labeling, which enables better aggregating crowd sequential annotations. But these approaches measure the quality of crowd labels under the assumption of only one ground-truth. As a result, the disagreement between annotators has to be considered as noisy labelings. However, research in NLP field shows that inter-disagreement among experts could be a good indicator for ambiguity and uncertainty that is inherent in language \cite{re8}. Apparently, there is no clear answer for the linguistically hard cases. As shown in Figure 1, ``like” can be tagged as conjunction or adjective. Furthermore, inter-disagreement between experts could reveal confusing label information that is related to the distribution of hard cases over label pairs. Figure 2 demonstrates label confusion matrix in POS tagging task, where ``ADJ" (adjectives) and ``NOUN" (nouns) are more likely to be confused. Wisely incorporating confusing label information into supervised learning can make the classifier more robust \cite{re8}. However, existing crowd sequential models do not take inter-disagreement between annotators into account.

\begin{figure}
\centering
\includegraphics[width=3.3in,height = 1.3 in ]{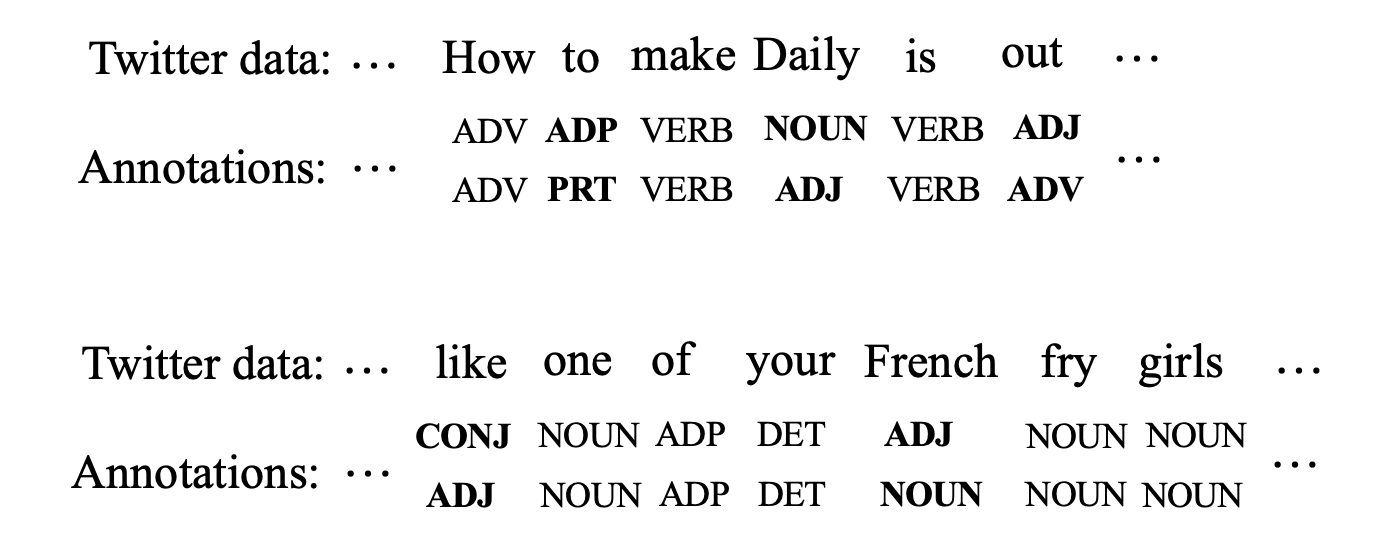}
\caption {Two examples of doubly-annotated twitter POS tagging data by different experts.}
\label{fig:secondfigure}
\end{figure}

\begin{figure}
\centering
\includegraphics[width=3in,height = 1.8in ]{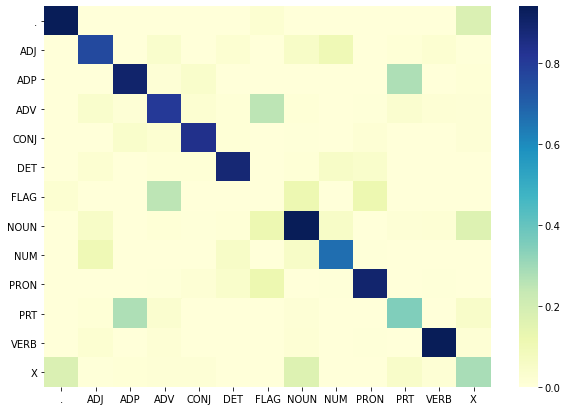}
\caption {Label confusion matrix derived from two gold annotations of 500 twitters POS tagging.}
\label{fig:secondfigure}
\end{figure}

To explore the inter-disagreement between reliable annotators and effectively preserve confusing label information, in this paper, we propose a framework called Learning Ambiguity from Crowd Sequential Annotations (LA-SCA). Our contributions can be summarized as follows:

First, we develop a hierarchical Bayesian model to group annotators into different clusters. By imposing an hierarchical prior on the confusion matrix that describes the reliability of annotators in the same cluster, the hierarchical Bayesian model allows the annotators that belong to the same cluster to be characterized with different but similar reliability, which aims to preserve inter-disagreement between reliable annotators.

Second, a low-rank model is formulated to model the relationship between the size of group the annotator involved in, annotator’s reliability and element's unambiguity in each sequence. Then inter-disagreement between reliable annotators on ambiguous elements can be obtained to compute label confusion matrix.

Third, cost-sensitive mechanism is combined to sequence labeling to encourage two more confusing label sequences that contain the ground-truth incur a lower Hamming loss, which aims to improve the robustness of sequence model.

\section{Related Work}

Hidden Markov Models (HMMs) \cite{re9,re10} and Conditional Random Fields (CRFs) \cite{re12,re16} form the most popular generative-discriminative pair for sequence labeling. With the great success of DL models, the combination of deep learning and graphical models receives increasing attention \cite{re17}. For example, Bi-LSTM-CRF \cite{re1} is proposed to efficiently encode past and future input features by combining a bidirectional LSTM network and a CRF layer. Furthermore, Transformer \cite{re2} is proposed with attention mechanism to learn long-range dependencies, which demonstrates significant improvement in efficiency and performance. However, both traditional and DL models require a large amount of training data with exact annotations, which is financially expensive and prohibitively time-consuming. Incorporating semi-supervised learning to sequence labeling models (e.g. semi-supervised CRFs and semi-SVM) can partly lighten the burden of sequential annotations, but this learning mechanism still needs exact labelings. 

Crowdsourcing provides an effective tool to collect large labeled dataset. Existing crowdsourcing learning models can be grouped into two types: wrapper and joint models. The former uses the inferred ground-truths from crowds for subsequent classifier learning while joint models simultaneously estimate annotators' reliability and learn the prediction model. Dawid \& Skene (DS) \cite{re18} aggregation model and its variants (e.g. GLAD \cite{re19}) explore different ways to model the relationship between the ground-truth, annotators' reliability and corresponding annotations, and then use Expectation Maximization (EM) approach to estimate the ground-truth labels and annotators' reliability. Sembler \cite{re7} and HMM-Crowd \cite{re6} models are proposed to aggregate multiple annotations to learn sequence model. \citet{re20} further took label dependency in sequential annotation into consideration and used a Bayesian approach to model crowd sequential labels. It should be addressed that the above crowdsourcing learning models assume only one ground-truth and do not consider the inter-disagreement among annotators. As a result, these models fail to capture inherent semantic ambiguity in NLP tasks and preserve confusing label information.

To explore inter-disagreement between annotators, \citet{re23} derived a label confusion matrix from doubly gold annotations and showed that the POS tagging classifier sensitive to confusing label information is more robust. \citet{re21} proposed CrowdTruth methodology to model ambiguity in semantic interpretation and treat all reliable annotator's labelings on ambiguous cases as high quality annotations. \citet{re22} used ambiguity-aware ground-truths to train the classifier for open-domain relation extraction and the results showed that ambiguity-aware crowds are better than the experts regarding the quality and efficiency of annotation. However, CrowdTruth based models only preserve multiple ground-truths for ambiguous instances and ignore confusing label information that can benefit robust classifier learning. 

In recent years multi-label crowdsourcing has been developed to identify multiple true labels from crowds for multi-label tasks. Different from discovering inter-disagreement between annotators in single-label crowdsourcing, most multi-label crowdsourcing methods assume that multiple ground-truths are assigned by one annotator. For example, \citet{re31} extended generative single-label crowdsourcing method by combining the correlation among labels while \citet{re30} further utilized neighbors’ annotation and effort-saving annotating behavior of each annotator to jointly estimate annotators' expertise and multi-label classifier. There have also been some research works exploring multi-label crowd consensus \cite{re32, re33} with the assumption that reliable annotators share the same label correlations, which fails to preserve inter-disagreement among reliable annotators, though.

\section{Proposed framework}
The proposed framework LA-SCA contains three parts: infer ground-truths and reliable annotators by hierarchical modeling of crowds; obtain confusing label information from inter-disagreement between reliable annotators on ambiguous elements via a low rank model; incorporate label confusion information in cost-sensitive sequence labeling. The details are described as follows:

\subsection{Hierarchical Modeling for Crowd Annotations}
Let $Y =\left\{y_{i1},y_{i2},...,y_{iL}\right\}_{i=1}^{N}$ denotes the crowd annotations provided by $L$ annotators over $N$ instances. Each annotator $l$ is belonged to a cluster $c\in \left \{ 1,2,...C \right \}$ and characterized by a confusion matrix $\Psi ^{c_l}\in[0,1]^{T\times T}$ where $T$ denotes the size of possible label set for $Y$.

We assume that the annotators in the same cluster have similar reliability but the corresponding annotations could be different. For example, annotators with lower reliability will provide various annotations in labeling a specific instance while reliable annotators have different opinions on ambiguous instances. To preserve disagreement between the annotators in the same cluster, we use the following hierarchical prior on each row of the confusion matrix $\Psi ^{c_l}$:
\begin{equation}
\eta^{c}_t \sim \mathrm{Exponential}(\lambda_t), \eta^{c}_t >0,
\end{equation}
\begin{equation}
\bm{\beta}_t^{c}\sim \mathrm{Dirichlet}({\bm{\alpha}_t}), 0< \bm{\beta}_{tj}^{c} <1, \sum_{j=1}^{T}\bm{\beta}_{tj}^{c} = 1,
\end{equation}
\begin{equation}
\Psi _t^{c_l}\sim \mathrm{Dirichlet}(\eta^{c}_t\bm{\beta}_t^{c}),
\end{equation}
\begin{equation}
y_{il}\sim \mathrm{Multinomial}(\Psi _{z_i}^{c_l}),
\end{equation}
where $c_l$ denotes that annotator $l$ belongs to cluster $c$ and $z_i$ is the ground truth label of $i_{th}$ instance. $\eta^{c}_t $ and $\bm{\beta}_t^{c}$ can be understood as the precision and mean of $\Psi _t^{c_l}$ respectively.

Besides, the cluster assignment $c_l$ and the ground truth $z_i$ follows multinomial distribution as follows:
\begin{equation}
c_l\sim \mathrm{Multinomial}(\bm{\nu}),
\end{equation}
\begin{equation}
z_i\sim \mathrm{Multinomial}(\bm{\gamma}),
\end{equation}
where $\bm{\nu}$ and $\bm{\gamma}$ are sampled from $\mathrm{Dirichlet}(\bm{\epsilon_{{\nu}}})$ and $\mathrm{Dirichlet}(\bm{\epsilon_{\gamma}})$ respectively.

We employ collapsed Gibbs sampling \cite{re24, re34} to estimate the conditional distribution over hidden variables $z_i$ and $c_l$ (more computation details can be found in Technical Appendix A). Let $\bm{c}$ and $\bm{z}$ denote the cluster assignments and true labels respectively, $\bm{c}^{-l}$ indicates that annotator $l$ is excluded from the cluster assignment and $\bm{z}^{-i}$ excludes $i_{th}$ instance. 
The conditional distribution of cluster assignment of annotator $l$ given the rest variables is computed as:
\begin{equation}
\begin{aligned}
&p(c_l=c|\bm{c}^{-l},\bm{z},Y,\eta,\bm{\beta })\\
&\propto p(c_l=c|\bm{c}^{-l})\times p(Y^{l}|Y^{-l},\bm{z},\bm{c},\eta,\bm{\beta})\\
&\propto (n_c^{-l}+\bm{\epsilon _\upsilon}/C )\times  \prod_{t}\frac{\Gamma (\eta _{t}^{c})}{\Gamma (n_{lt}+\eta _{t}^{c})}\prod_{s}\frac{\Gamma (n_{lts}+\eta _{t}^{c}\bm{\beta }_{ts}^{c})}{\Gamma (\eta _{t}^{c}\bm{\beta }_{ts}^{c})},\end{aligned}
\end{equation}
where $n_c^{-l}$ denotes the number of annotators (exclude $l$) assigned to cluster $c$. $n_{lt}$ is the number of instances that are annotated by $l$ and have true label $t$. $n_{lts}$ denotes the number of instances that are annotated with label $s$ by $l$ and have true label $t$.

Similarly, $p(z_i=t|\bm{z}^{-i},Y,\bm{c},\eta,\bm{\beta })$ is given as
\begin{equation}
\begin{aligned}
&p(z_i=t|\bm{z}^{-i},Y,\bm{c},\eta,\bm{\beta} )\\
&\propto p(z_i=t|\bm{z}^{-i})\times p(y_{i}|Y^{-i},z_i = t,\bm{z}^{-i},\bm{c},\eta,\bm{\beta})\\
&\propto (n_t^{-i}+\bm{\epsilon _\gamma} /T)\times \prod_{c}\prod_{c_l=c}\frac{\prod_{s}(n_{lts}^{-i}+\eta _{t}^{c}\bm{\beta} _{ts}^{c})^{I(y_{il}=s)}}{n_{lt}^{-i}+\eta _{t}^{c}},
\end{aligned}
\end{equation}
where $n_{t}^{i}$ denotes the number of instances (exclude $i$) with true label $t$. $n_{lt}^{i}$ denotes the number of instances (exclude $i$) that are annotated by $l$ and have true label $t$. $n_{lts}$ denotes the number of instances (exclude $i$) that are annotated with label $s$ by $l$ and have true label $t$.

Due to non-conjugacy of Exponential and Dirichlet prior for the likelihood function $p(Y|\bm{z},\bm{c},\eta,\bm{\beta})$, we use Metropolis-Hastings (MH) algorithm \cite{re25} to estimate the conditional posterior distribution $p(\bm{\beta}_{tj}^c|\bm{\beta}_{t(\sim j)}^c,\eta _t^c,Y^c)$ and $p(\eta _t^c|\bm{\beta}_t^c ,Y^c)$ for each cluster, and the symmetric proposal distribution (i.e. uniform distribution) is selected to simulate a candidate sample (algorithm details are presented in Technical Appendix C Algorithm 1 and 2).

$p(\bm{\beta}_{tj}^c|\bm{\beta}_{t(\sim j)}^c,\eta _t^c,Y^c)$ is given as 
\begin{equation}
\begin{aligned}
&p(\bm{\beta}_{tj}^c|\bm{\beta}_{t(\sim j)}^c,\eta _t^c,Y^c)\propto p(Y^{cj}|Y^{-cj},\bm{\beta}_{t}^c,\eta _t^c)\times p(\bm{\beta}_{tj}^c|\bm{\beta}_{t(\sim j)}^c)\\
&\propto \prod_{c_l= c}\left \{ \frac{\Gamma (n_{ltj}+\eta _t^c\bm{\beta}_{tj}^c)}{\Gamma (\eta _t^c\bm{\beta}_{tj}^c)} \frac{\Gamma (n_{ltt}+\eta _t^c\bm{\beta}_{tt}^c)}{\Gamma (\eta _t^c\bm{\beta}_{tt}^c)}\right \}\\&\times \left ( \frac{\bm{\beta}_{tj}^c}{1-u_{tj}^c} \right )^{\lambda_t \bm{\alpha} _{tj}-1}\left ( 1- \frac{\bm{\beta}_{tj}^c}{1-u_{tj}^c}\right )^{\lambda_t \bm{\alpha} _{tt}-1},
\end{aligned}
\end{equation}
where $\bm{\beta}_{t}^c=\left [\bm{ \beta}_{t1}^c ,...,\bm{\beta}_{tj}^c,...,\bm{\beta}_{tT}^c\right ]$. $u_{tj}^c=1-\sum_{s=1,s\neq t,s\neq j} ^{T}\bm{\beta}_{ts}^{c}$. The conditional posterior distribution of $\bm{\beta}_{tt}^c$ is obtained via $\bm{\beta}_{tt}^c=1-\sum_{j=1,j\neq t}^T \bm{\beta}_{tj}^{c}$. Derivation details of $p(\bm{\beta}_{tj}^c|\bm{\beta}_{t(\sim j)}^c,\eta _t^c,Y^c)$ can be found in Technical Appendix B.

$p(\eta _t^c|\bm{\beta}_t^c ,Y^c)$ is defined as
\begin{equation}
\begin{aligned}
p(\eta _t^c|\bm{\beta}_t^c ,Y^c)\propto& \prod_{c_l= c}\frac{\Gamma(\eta _t^c)}{\Gamma (n_{lt}+\eta _t^c)}\\
&\times\prod_{j}\frac{\Gamma(n_{ltj}+\eta _t^c\bm{\beta} _{tj}^c)}{\Gamma (\eta _t^c\bm{\beta} _{t}^c)}\times \lambda_t e^{-(\lambda_t\eta _t^c)}.
\end{aligned}
\end{equation}

\cite{re35}

By iteratively estimating $z_i$, $c_l$, $\bm{\beta}_{t}^c$ and $\eta_t^c$ until convergence, annotators that have similar reliability could be grouped into the same cluster. Since inter-disagreement among experts could reveal linguistically ambiguous cases, to identify the cluster with reliable annotators, we compute the shared confusion matrix of each cluster based on the estimated ground truths $\bm{z}$. One ground-truth for each instance does not affect estimation of the shared confusion matrix as ambiguous cases only take up a very small part in the whole dataset \cite{re23}. The entry of the shared confusion matrix $M^c \in \mathcal{R}^{T \times T}$ for the cluster $c$ is defined as
\begin{equation}
M^c_{t,s}=\frac{\sum_{i}I(z_i=t)\prod\nolimits_{{l\in c}}I(y_{il}=s)}{\sum_{i}I(z_i=t)},
\end{equation}
and the high-reliability cluster is obtained with $\arg\max_c\sum_{t=1}^{T}M^c_{t,t} \Big /T$.

\subsection{Identifying Ambiguity via Low Rank Model} 

Based on the identified reliable annotators, to estimate ambiguity degree of each element in a sequence, we assume that in the high-reliability cluster the decisions of annotators form small groups for ambiguous elements and annotators who is more reliable in labeling this sequence is consistent with other annotators for unambiguous elements. Inspired by the quantitative formula used to describe the relationship between the size of group, annotator's reliability and task clarity \cite{re26}, we construct an $L\times N^s$ matrix $A$ for $s_{th}$ sequence where $N^s$ is the length of the sequence. In this matrix, each entry $A(l,N_j^s)$ denotes the size of group for annotator $l$ involved in labeling $j_{th}$ element in $s_{th}$ sequence. We define $A(l,N_j^s)$ as
\begin{equation}
A(l,N_j^s)=\omega_l^s\times \mu _j^s,
\end{equation}
where $\omega_l^s$ represents the reliability of annotator $l$ in labeling $s_{th}$ sequence and  $\mu _j^s$ is the degree of unambiguity of $j_{th}$ element.

Intuitively, if the annotator is more reliable or the element is less ambiguous, the size of group is more larger. Thus we employ rank-1 factorization to formulate the relationship between $A(l,N_j^s)$, $\omega_l^s$ and $\mu _j^s$. The degree of unambiguity of each element in $s_{th}$ sequence is computed as follows:

\begin{equation}
A^s=U\Lambda V^T,
\end{equation}

\begin{equation}
\bm{\omega}^s=U_{.1}\sqrt{\Lambda _{11}},
\end{equation}

\begin{equation}
\bm{\mu}^s=V_{.1}\sqrt{\Lambda _{11}},
\end{equation}
where $\bm{\omega}^s=\left [ \omega^s_1,\omega^s_2,...,\omega^s_L \right ]$ and $\bm{\mu}^s=\left [ \mu_1,\mu_2,...,\mu_{N_s} \right ]$.

There are three steps concerning identifying ambiguity:

\textbf{a. identify ambiguous elements.} We rank the set of estimated degree of unambiguity for the whole sequential data and choose an appropriate percentage $p$ to identify the element that falls in the range of top $p$ minimum as ambiguous cases. 

\textbf{b. compute inter-disagreement between annotators.} For the identified ambiguous cases, the disagreement among reliable annotators provides multiple possible ground-truths. Let $\left \{ y_{jt} \right \}_{t=1}^{t=L'}(L'<= L)$ denotes the set of labels assigned by annotators for $j_{th}$ ambiguous elements in $s_{th}$ sequence, the score of $y_{jt}$ can be defined as
\begin{equation}
S(y_{jt})=\frac{\sum_{l=1}^{L}I(y_{jl}=y_{jt})\omega^s_l}{\sum_{l=1}^{L}I(y_{jl}=y_{jt})}.
\end{equation}

In practice ambiguous instances have limited gold annotations \cite{re8}. We select top two labels for each ambiguous element by $S(y_{jt})$ in descending order and combine them with the inferred ground-truth in hierarchical modeling. 

\textbf{c. obtain confusing label information.} Label confusion matrix $CF\in \mathcal{R}^{T \times T}$ is utilized to show the degree of confusion between label pairs, and the entry $CF(i,j)$ is defined as the mean of $p(z(x)=i,z(x)=j)$ and $p(z(x)=j,z(x)=i)$ where $p(z(x)=i,z(x)=j)$ is computed as
\begin{equation}
p(z(x)=i,z(x)=j)=\frac{\sum_{k}I(z(x_k)=i,z(x_k)=j)}{\sum_{k}I(z(x_k)=i)},
\end{equation}
where $k$ denotes $k_{th}$ element in the whole sequential dataset, and $p(z(x)=j,z(x)=i)$ is computed in a similar way.

\subsection{Cost-sensitive sequence labeling}

Given $\left \{\bm{x}_i,\bm{z}_i \right \}_{i=1}^{N}$ sequential dataset, where $\bm{z}_i$ are the inferred ground-truths via hierarchical Bayesian modeling. Traditional training criteria is to maximize the likelihood of conditional log-linear model, which does not distinguish the ground-truth from all incorrect outputs that are penalized equally through normalization. To improve sequence labeling, we employ cost-sensitive mechanism to incorporate confusing label information in the training, where the label sequence that is more confusing with the ground-truth incurs lower cost. The objective of cost-sensitive sequence labeling is defined as

\begin{equation}
L_{\mathrm{CS}}(\bm{\theta}) =\sum_{i=1}^{N}\log \frac{\exp\left \{\bm{\theta}^T\bm{f}(\bm{x}_i,\bm{z}_i)\right \}}{\sum\limits_{\bm{z}_j}\mathrm{cost}(\bm{z}_j,\bm{z}_i)\exp \left\{
\bm{\theta}^T\bm{f}(\bm{x}_i,\bm{z}_j)\right \} },
\end{equation}
where $\bm{f}(\bm{x}_i,\bm{z}_i)$ denotes the feature function. $\mathrm{cost}(\bm{z}_j,\bm{z}_i)$ is used to measure the influence of confusing label information on the loss. A weighted Hamming loss is defined to describe $\mathrm{cost}(\bm{z}_j,\bm{z}_i)$  as
\begin{equation}
\mathrm{cost}(\bm{z}_j,\bm{z}_i)=\frac{1}{K_i}\sum_{k=1}^{K_i}\left ( 1-p(\bm{z}_{jk},\bm{z}_{ik})\right )*\left ( \bm{z}_{jk}\oplus \bm{z}_{ik}\right ),
\end{equation}
where $K_i$ is the number of tokens in $i_{th}$ sequence. $\oplus$ is the XOR boolean operator, $p(\bm{z}_{jk},\bm{z}_{ik})$ is obtained from label confusion matrix.


\section{Experiments}
We conduct experiments on POS tagging and NER for English. It is widely debatable of POS analysis where there are many hard cases that annotators disagree on \cite{re8}, while in NER the definition and partition of named entity still remains arguable. In the following sections, we present quantitative results to investigate the effectiveness of our framework in inferring the ground-truths, predicting unknown sequences and preserving confusing label information.

\subsection{Datasets}
Current published datasets cannot satisfy both crowd annotations and multiple gold annotations. We employ multiple gold-annotated and crowd-annotated datasets as follows:

 \textbf{POS tagging}: Most POS tagging datasets only contain one gold annotation which fail to identify hard cases. Therefore we use three twitter POS tagging datasets in the work of studying cost-sensitive POS tagger \cite{re22}, which include 500 tweets with doubly gold annotations ( denoted as T-DGA for simplicity),  RITTER-TEST (118 tweets) dataset and INHOUSE (200 tweets) dataset. We employ T-DGA as training data (doubly gold annotations guarantee the existence of hard cases), and RITTER-TEST and INHOUSE as test datasets\footnote{Both RITTER-TEST and INHOUSE have only one gold annotation.}.
  
\textbf {NER}: CoNLL-2003 shared NER dataset \cite{re27} is one of the most common benchmarks used in NLP community for sequence labeling, which contains four types of entities: persons (PER), locations (LOC), organizations (ORG) and miscellaneous (MISC). Rodrigues et al. \cite{re28} put 400 articles from CoNLL-2003 on Amazon’s Mechanical Turk to collect crowd annotations. There are total 47 annotators and the average number of annotators per article is 4.9. In this paper, after pre-processing these crowd-labeled data we select 3000 sentence-level sequences, and use CoNLL 2003 test data\footnote{CoNLL 2003 testset has only one gold annotation.}.

\subsection{Baselines}

We use the following six models to learn from crowd sequential data as baselines.

MVtoken \cite{re27}: The ground-truth label sequence is obtained by choosing the label with more votes in token level.

DS \cite{re18}: The EM algorithm is employed to assign weight to each vote in token level.

MACE \cite{re29}: By including a binary latent variable that denotes if and when each annotator is spamming, the model can identify which annotators are trustworthy and produce the true label.

Sembler \cite{re7}: The model extends crowdcoursing learning on instance level to sequence level and jointly estimate annotators' reliability and sequence model.

HMM-Crowd \cite{re6}: Based on HMMs, the model further models the “crowd component” by including the parameters for the label quality of annotators and crowd variables.

HC-CLL: To verify the effectiveness of cost-sensitive sequence labeling, we also train the sequence prediction model by maximizing conditional log-likelihood.

\subsection{Experimental setting}

Synthetic crowd annotations: As T-DGA does not have real crowd annotations, we simulate annotators with different reliability by controlling the precision of their annotations. In practice the number of annotators is limited, we set the total number of annotators as 15 and arrange three different assignments: $[5,5,5]$, $[8,4,3]$ and $[3,4,8]$. In each assignment, three different ranges of precision: $[0.9,0.7]$, $[0.7,0.4]$ and $[0.4,0.1]$ are set to indicate various reliability from high to low levels.

LA-SCA framework: The optimal number of clusters for annotators is selected between the range $[2,5]$ based on Bayesian information criteria. $\lambda_t$ is set to 2. To confirm that crowd annotations are better than randomly labeling, the diagonal of $\bm{\alpha}_t$ is set to 0.7 while the off diagonal elements are set to 0.3. Furthermore, we select $p = 10\%$ to identify ambiguous elements.

\subsection{Experimental results}

\subsubsection{Comparing with baselines}
We evaluate the effectiveness of the proposed framework in inferring ground-truths for training data and predicting testset.

\textbf{POS tagging task:}

For simplicity, we denote three different crowd annotations $[8,4,3]$, $[5,5,5]$ and $[3,4,8]$ as $\mathrm{ca}_1$, $\mathrm{ca}_2$ and $\mathrm{ca}_3$, respectively. Table 1 shows accuracy of inferring ground-truths in T-DGA dataset (HC-CLL is the same as LA-SCA in inferring ground-truths). We can see that most of crowd models achieve better performance by increasing the proportion of high quality annotations. The performance of each comparing model varies in Gold 1 and Gold 2 as these two gold annotations have different label assignments for some tokens. For the case of low quality annotations (i.e. $\mathrm{ca}_3$), the developed hierarchical Bayesian model effectively identifies the annotators with high reliability, which can help guide the estimation of ground-truths and thus improves the performance. DS and HMM-Crowd achieve competitive results as the mechanism of iteratively estimate annotators' reliability and the ground-truths alleviates the negative effect of low quality annotations. 
\begin{table}[H]
\centering
\caption{Accuracy of inferring ground-truths for T-DGA dataset (\%).}
\begin{adjustbox}{width=0.48\textwidth}
\begin{tabular}{lcccccc} 
\hline\addlinespace[0.1cm]
\multirow{2}{*}{~Model} & \multicolumn{3}{c}{Gold1}                        & \multicolumn{3}{c}{Gold2}                         \\
                        & ca1            & ca2            & ca3            & ca1            & ca2            & ca3             \\ 
\hline\addlinespace[0.1cm]
MVtoken                 & 91.82          & 90.84          & 83.70          & 81.94          & 81.81          & 73.20           \\\addlinespace[0.1cm]
DS                      & 93.34          & 90.97          & 91.99          & 83.53          & \textbf{81.90} & 82.14           \\\addlinespace[0.1cm]
MACE                    & 89.80          & 84.79          & 84.56          & 80.76          & 76.26          & 75.89           \\\addlinespace[0.1cm]
Sembler                 & 93.05          & 89.58          & 85.78          & 83.36          & 80.22          & 76.99           \\\addlinespace[0.1cm]
HMM-Crowd               & \textbf{93.40} & 91.59          & 90.38          & \textbf{83.72} & 82.06          & 81.12           \\\addlinespace[0.1cm]
LA-SCA                  & 93.30          & \textbf{92.59} & \textbf{92.22} & 83.47          & 81.73          & \textbf{83.71}  \\
\hline
\end{tabular}
\end{adjustbox}
\end{table}

Table 2 reports F1 score of comparing methods on RITTER-TEST and INHOUSE datasets. Generally the model that learns from higher quality of ground-truths can achieve better prediction performance. For the wrapper models that input the inferred ground-truths to the sequence model (i.e. MVtoken, DS and MACE), prediction performance heavily depends on the quality of the estimated ground-truths. Therefore in $\mathrm{ca}_3$ setting, the F1 score of wrapper models (i.e. MVtoken, DS and MACE) is lower than that of joint models (i.e. Sembler and HMM-Crowd). The developed hierarchical Bayesian model HC-CLL effectively identifies the cluster with high reliability which enables stable performance in handling low quality annotations. Compared with HC-CLL, LA-SCA achieves better results in $\mathrm{ca}_1$ and $\mathrm{ca}_2$ settings as low quality crowd annotations (i.e. $\mathrm{ca}_3$) fail to provide effective confusing label information, which is more likely to add much noise in cost-sensitive sequence labeling and then degrades prediction performance.

\begin{table}[H]
\centering
\caption{Performance of the models on RITTER-TEST and INHOUSE dataset (\%).}
\begin{adjustbox}{width=0.48\textwidth}
\begin{tabular}{lcccccc} 
\hline\addlinespace[0.1cm]
\multirow{2}{*}{Model} & \multicolumn{3}{c}{ RITTER-
TEST } & \multicolumn{3}{c}{INHOUSE}                              \\
                       & $\mathrm{ca}_1$ & $\mathrm{ca}_2$ &   $\mathrm{ca}_3 $           &  $ \mathrm{ca}_1$            &  $\mathrm{ca}_2$              &  $\mathrm{ca}_3  $             \\ 
\hline\addlinespace[0.1cm]
MVtoken                &59.35  & 58.72 &   58.72             & 53.15          & 52.29          & 48.03           \\\addlinespace[0.1cm]
DS                     & \textbf{67.58} & 60.43          & 58.69                  & 54.33          & 48.49          & 48.14           \\\addlinespace[0.1cm]
MACE                  & 59.89          & 58.04          & 60.05               & 47.92          & 53.33          & 49.12           \\\addlinespace[0.1cm]
Sembler               & 59.82          & 61.74          & 60.53             & 49.65          & 50.22          & 49.93           \\\addlinespace[0.1cm]
HMM-Crowd              & 61.10          & 61.30          & 61.79                & 49.98          & 49.12          & 52.40           \\\addlinespace[0.1cm]
HC-CLL                  & 61.18          & 65.12          & \textbf{62.57}               & 52.97          & 54.55          & \textbf{52.88}  \\\addlinespace[0.1cm]
LA-SCA                 & 66.20          & \textbf{67.32} & 61.57                & \textbf{55.65} & \textbf{57.25} & 50.84           \\
\hline
\end{tabular}
\end{adjustbox}
\end{table}

\textbf{NER task:}

In NER tagging task class ``O" accounts for a great proportion of the total classes, thus we use F1 score instead of accuracy to report the performance of inferring ground-truths for training data of CoNLL 2003 NER task. As shown in Table 3, the developed hierarchical Bayesian model achieves the best F1 score and DS model also achieves competitive result. Table 3 also demonstrates the performance of predicting labels for testing data. Due to limited crowded training data the overall performance of comparing methods is well below the reported results \cite{re28}. The proposed framework LA-SCA still outperforms the baselines but only by a narrow margin. Since cost-sensitive learning mechanism inevitably produces label noises in NER task as there are a few confusing labels that should be attended to each other, directly maximizing log-likelihood can be competitive with cost-sensitive maximization.

\begin{table}[H]
\centering
\caption{Evaluation on CoNLL 2003 NER task (\%).}
\begin{tabular}{lcc} 
\hline
Model     & Infer ground-truths &  Prediction \\ 
\hline
MVtoken   & 63.17 & 38.52  \\
DS        & 65.32 & 39.21  \\
MACE      & 60.07 & 37.10  \\
Sembler   &63.25  & 38.87  \\
HMM-Crowd &63.44  & 39.31  \\
HC-CLL    &\textbf{67.54}  & 40.56  \\
LA-SCA    & \textbf{67.54} & \textbf{41.56}  \\
\hline
\end{tabular}
\end{table}

\subsubsection{Identifying ambiguous cases}

In this section, we investigate the performance of LA-SCA in identifying ambiguous cases and preserving confusing label information. We present the results on T-DGA dataset ($\mathrm{ca}_1$ setting) as it provides the standard for comparison.

\begin{figure}[h]
\centering
\stackunder{\includegraphics[scale=0.4]{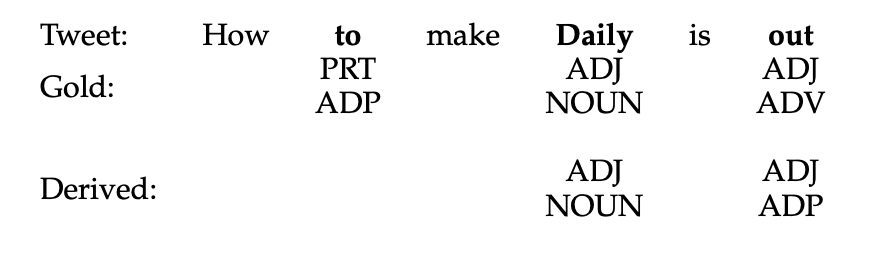}}{}
\par\bigskip
\stackunder{\includegraphics[scale=0.4]{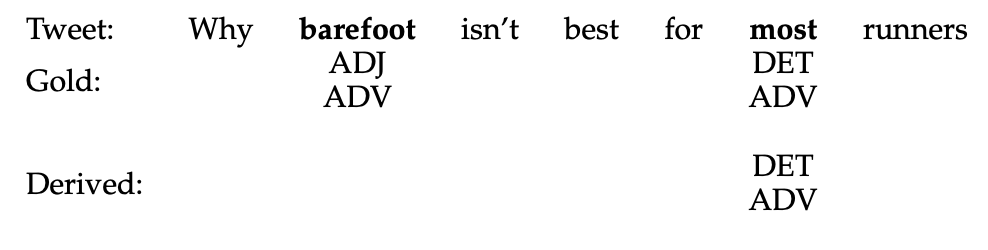}}{}
\caption{Two examples from T-DGA with gold and derived labelings on ambiguous cases.}
\end{figure}

First, we measure the performance of identifying ambiguous cases with the following measures:
\begin{equation}
\mathrm{acc}_1 = \frac{ \# \mathrm{correctly}\ \mathrm{identified}\ \mathrm{ambiguous}\ \mathrm{cases}}{\# \mathrm{all}\ \mathrm{ambiguous}\ \mathrm{cases}},
\end{equation}
\begin{equation}
\mathrm{acc}_2 = \frac{ \# \mathrm{correctly} \ \mathrm{double} \ \mathrm{annotated} \ \mathrm{ambiguous} \ \mathrm{cases}}{\# \mathrm{all}\ \mathrm{ambiguous}\ \mathrm{cases}},
\end{equation}
and we obtain that $\mathrm{acc}_1 = 725/931= 0.779$ and $\mathrm{acc}_2 = 614/931= 0.660$. It can be concluded that LA-SCA successfully identifies most of ambiguous cases in T-DGA. We further present two examples from T-DGA with gold and derived labelings on ambiguous cases, as demonstrated in Figure 3, LA-SCA identifies ambiguous cases with label confusing pairs of [``ADJ", ``NOUN"] and [``DET", ``ADV"] successfully.

\begin{figure}[h]
  \begin{subfigure}[b]{0.5\linewidth}
    \centering
    \includegraphics[width=1\linewidth]{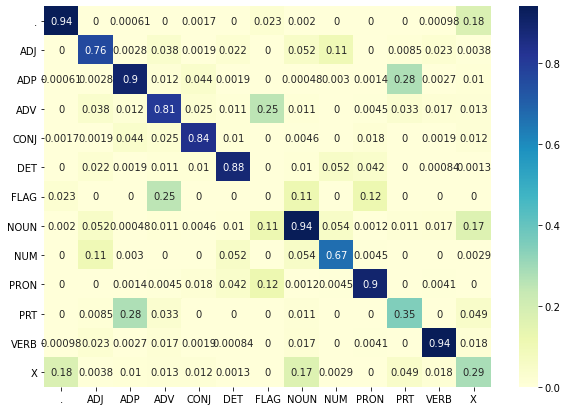} 
    \caption{Gold matrix.} 
    \label{fig7:a} 
    \vspace{4ex}
  \end{subfigure}
  \begin{subfigure}[b]{0.5\linewidth}
    \centering
    \includegraphics[width=1\linewidth]{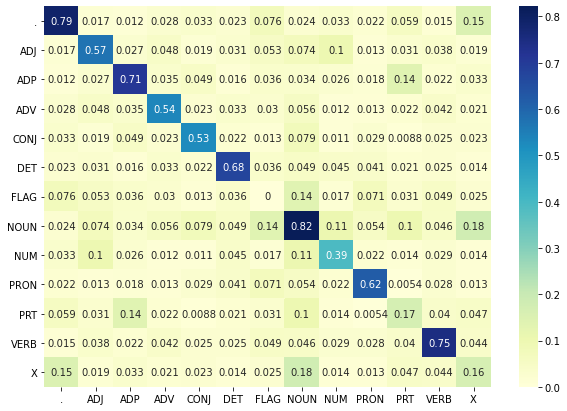} 
    \caption{Derived matrix ($\mathrm{ca}_1$).} 
    \label{fig7:b} 
    \vspace{4ex}
  \end{subfigure} 
  \begin{subfigure}[b]{0.5\linewidth}
    \centering
    \includegraphics[width=1\linewidth]{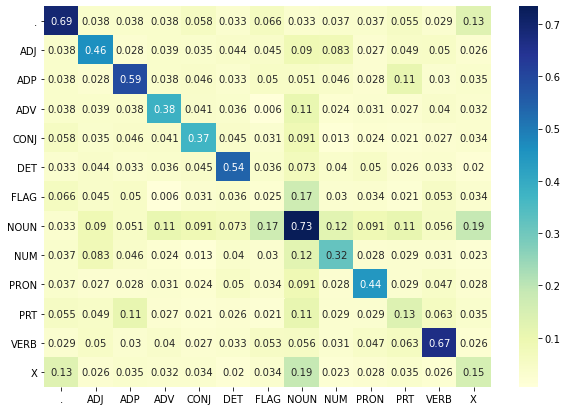} 
    \caption{Derived matrix ($\mathrm{ca}_2$).} 
    \label{fig7:c} 
  \end{subfigure}
  \begin{subfigure}[b]{0.5\linewidth}
    \centering
    \includegraphics[width=1\linewidth]{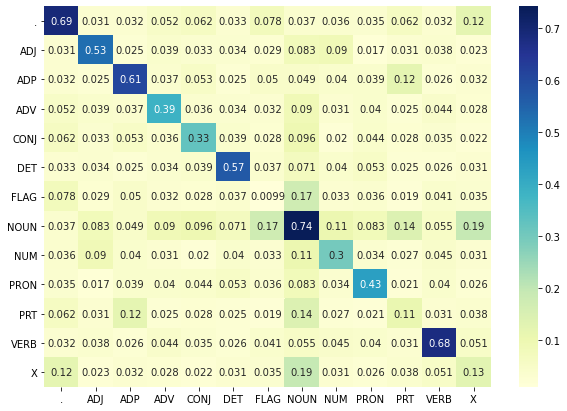} 
    \caption{Derived matrix ($\mathrm{ca}_3$).} 
    \label{fig7:d} 
  \end{subfigure} 
  \caption{Comparison between the gold the derived label confusion matrices on T-DGA.}
  \label{fig7} 
\end{figure}

\begin{figure*}
\minipage{0.3\textwidth}
  \includegraphics[width=0.95\linewidth]{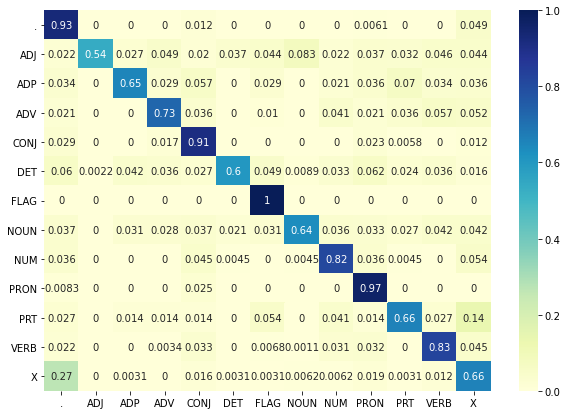}
  \subcaption{Cluster 1}\label{fig:awesome_image1}
\endminipage\hfill
\minipage{0.3\textwidth}
  \includegraphics[width=0.95\linewidth]{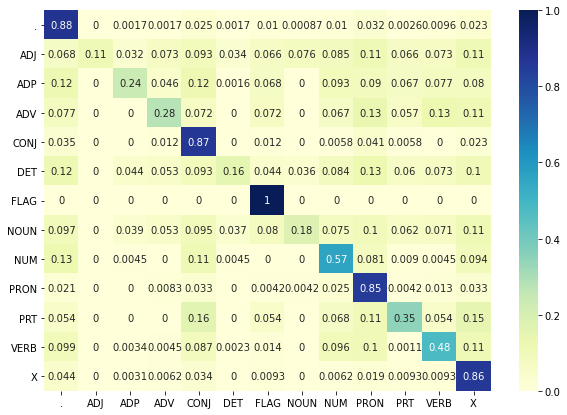}
  \subcaption{Cluster 2}\label{fig:awesome_image2}
\endminipage\hfill
\minipage{0.3\textwidth}%
  \includegraphics[width=0.95\linewidth]{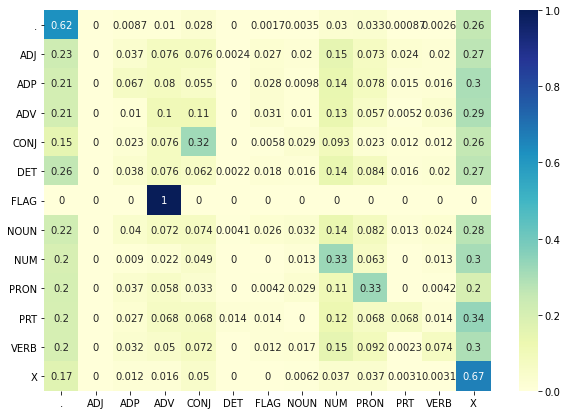}
  \subcaption{Cluster 3}\label{fig:awesome_image3}
\endminipage
\caption{Shared confusion matrices of three clusters on POS tagging task.}
\end{figure*}

Figure 4 shows the gold label confusion matrix and the derived confusion matrices of three settings respectively. As crowds contain noisy label information, the agreement for some labels is lower than the gold one, which may generate wrong confusing label information. For example, in the twitter ``FollowerSale is most Trusted company to buy”, the gold annotation of ``to” is  [``VERB”]  while the derived confusing label set is  [``VERB”, ``PRON”]. But the derived matrices also preserve some confusing label information that is similar to the gold. For example, the agreement between adjectives [``ADJ”] are nouns [``NOUN”], and [``X”] category is more likely to be confused with punctuations [``.”] and nouns [``NOUN”].

\subsubsection{Interpreting clusters}

Clustering for crowd annotations can help discover common patterns of the annotators with similar reliability. Here we demonstrate how to interpret the shared confusion matrices of the estimated clusters in POS tagging and NER tasks. 

\textbf{POS tagging task:}

We choose $\mathrm{ca}_1$ setting and present the shared confusion matrices of three clusters. By reviewing the diagonal elements in three shared confusion matrices of Figure 5, we can see that the developed Bayesian hierarchical model separates the annotators with different reliability well. Cluster 1 demonstrates the annotators with high reliability where the average successful identification value is above 0.7, while the annotators with lower reliability are clustered into the third cluster where the average successful identification value is below 0.3.

\textbf{NER task:}

We present clustering results of crowd annotations collected from AMT. Generally these crowd annotations are of good quality. It can be seen from Figure 6 that both two clusters are reliable in assigning ``I-PER", ``O", ``I-LOC" and ``B-ORG". Cluster 1 shows the more reliable annotations where the average diagonal value is 0.742, and the average diagonal value in cluster 2 is 0.535.

\begin{figure}[h]
  \begin{subfigure}[b]{0.5\linewidth}
    \centering
    \includegraphics[width=1\linewidth]{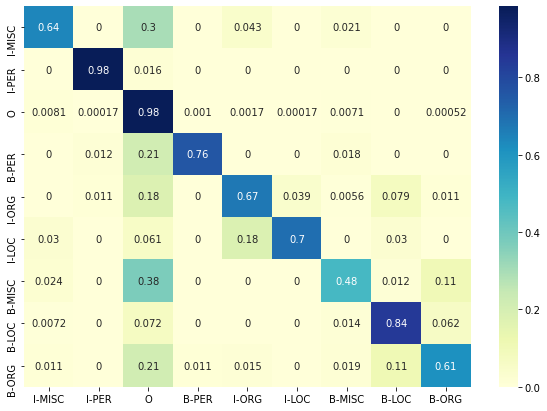} 
    \caption{Cluster 1.} 
    \label{fig7:a} 
    \vspace{4ex}
  \end{subfigure}
  \begin{subfigure}[b]{0.5\linewidth}
    \centering
    \includegraphics[width=1\linewidth]{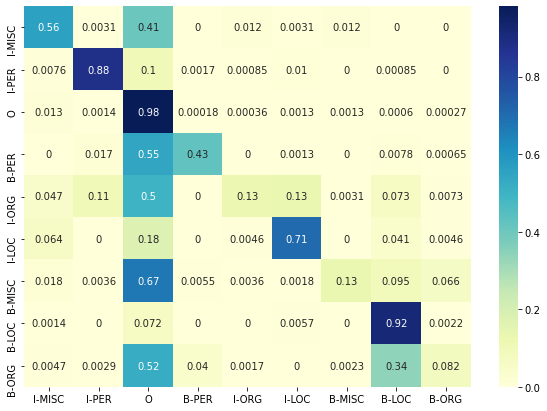} 
    \caption{Cluster 2.} 
    \label{fig7:b} 
    \vspace{4ex}
  \end{subfigure} 
    \caption{Shared confusion matrices of clusters on NER task.}
  \label{fig7} 
\end{figure}

\subsection{Discussion} 
The major concern regarding LA-SCA evaluation is to perform on the dataset with both crowd annotations and multiple gold annotations. For T-DGA, we have to simulate crowd annotations and use precision to indicate annotator's reliability in global view. A situation will arise that simple cases are more likely to be incorrectly annotated in the precision with $[0.9, 0.7]$, which is not in line with the decision of reliable annotators. In the crowd-annotated CoNLL-2003 NER task, the size of assigned annotators per article is limited and crowd annotations are of good quality, which more or less hinders exploration of labeling diversity. 

While the incomplete datasets partially limit the applicability of LA-SCA, the proposed hierarchical Bayesian modeling shows its competitiveness in inferring ground-truths from real crowd annotations and synthetic crowds with low reliability. As cost-sensitive mechanism expects sparse label confusion matrix in the task (e.g. NER) where only a few labels are more confused with each other, it still remains to be explored to achieve significant improvement in predicting unknown sequences.

\section{Conclusion}
In this paper, we propose a framework called Learning Ambiguity from Crowd Sequential Annotations (LA-SCA) to explore inter-disagreement between reliable annotators and effectively preserve confusing label information to improve robust sequence classifier learning. Experimental results show that LA-SCA achieves competitive performance in inferring ground-truth from crowds and predicting testset. Further, identified clusters can help interpret labeling patterns of  the annotators with similar reliability, which can help task designers improve labeling guideline.

\appendix

\section{A. Estimating the conditional distribution over hidden variables $z_i$ and $c_l$ }

a. Derivation of $p(c_l=c|\bm{c}^{-l},\bm{z},Y,\eta,\bm{\beta })$ is as follows:
\begin{equation} 
\scalebox{0.9}{$
p(c_l=c|\bm{c}^{-l},\bm{z},Y,\eta,\bm{\beta })\propto p(c_l=c|\bm{c}^{-l})\times p(Y^{l}|Y^{-l},\bm{z},\bm{c},\eta,\bm{\beta})$},
\end{equation} 
where $p(c_l=c|\bm{c}^{-l})$ is computed as
\begin{equation}
\begin{aligned}
&p(c_l=c|\bm{c}^{-l})\\
&=p(c_1,c_2,..,c_l=c)/p(c_1,c_2,...,c_{l-1})\\
&=\scalebox{0.95}{$\frac{\Gamma(\epsilon_\nu )}{\Gamma(L+\epsilon_\nu)}\prod_{c=1}^{C}\frac{\Gamma(n_c+\epsilon_\nu/C)}{\Gamma(\epsilon_\nu/C)}\Big/ \frac{\Gamma(\epsilon_\nu )}{\Gamma(L+\epsilon_\nu-1)}\prod_{c=1}^{C}\frac{\Gamma(n_{c}^{-l}+\epsilon_\nu/C)}{\Gamma(\epsilon_\nu/C)}$}\\
&=\frac{\Gamma(L+\epsilon_\nu-1)}{\Gamma(L+\epsilon_\nu)}\frac{\Gamma(n_{c}+\epsilon_\nu/C)}{\Gamma(n_{c}^{-l}+\epsilon_\nu/C)}\\
&=\frac{n_{c}^{-l}+\epsilon_\nu/C}{L+\epsilon_\nu-1}.
\end{aligned}
\end{equation}

The likelihood $p(Y|\bm{z},\bm{c},\eta,\bm{\beta}$ is obtained by integrating out the variables $\Psi ^{c_l}$:

\begin{equation} \tag{3}
\begin{aligned}
&p(Y|\bm{z},\bm{c},\eta,\bm{\beta})\\
&=\prod_{c}\prod_{l\in c}\prod_t\int p(Y^{c_l}|\Psi _t^{c_l})p(\Psi _t^{c_l}|\eta^{c}\bm{\beta}_t^c)d(\eta^{c}\bm{\beta}_t^c)\\
&=\prod_{c}\prod_{l\in c}\prod_t\left [ \frac{\Gamma (\eta _{t}^{c})}{\Gamma (n_{lt}+\eta _{t}^{c})}\prod_{s}\frac{\Gamma (n_{lts}+\eta _{t}^{c}\bm{\beta} _{ts}^{c})}{\Gamma (\eta _{t}^{c}\bm{\beta}_{ts}^{c})}\right ].
\end{aligned}
\end{equation}

Then via Bayesian rule we obtain the conditional distribution:
\begin{equation} \tag{4}
\scalebox{0.9}{$
 p(Y^{l}|Y^{-l},\bm{z},\bm{c},\eta,\bm{\beta}) = \prod_t\left [ \frac{\Gamma (\eta _{t}^{c})}{\Gamma (n_{lt}+\eta _{t}^{c})}\prod_{s}\frac{\Gamma (n_{lts}+\eta _{t}^{c}\bm{\beta }_{ts}^{c})}{\Gamma (\eta _{t}^{c}\bm{\beta}_{ts}^{c})}\right ]$}.
\end{equation}

b. Derivation of $p(z_i=t|\bm{z}^{-i},Y,\bm{c},\eta,\bm{\beta })$ is as follows:
\begin{equation} \tag{5}
\begin{aligned}
&p(z_i=t|\bm{z}^{-i},Y,\bm{c},\eta,\bm{\beta} )\\
&\propto p(z_i= t|\bm{z}^{-i})\times p(y_{i}|Y^{-i},z_i = t, \bm{z}^{-i},\bm{c},\eta,\bm{\beta}),
\end{aligned}
\end{equation}
where $p(z_i|\bm{z}^{-i})$ is computed as
\begin{equation} \tag{6}
\begin{aligned}
&p(z_i=t|\bm{z}^{-i})\\
&=p(z_1,z_2,..,z_i=t)/p(z_1,z_2,...,z_{i-1})\\
&= \scalebox{0.95}{$
\frac{\Gamma(\epsilon_\gamma)}{\Gamma(N+\epsilon_\gamma )}\prod_{t=1}^{T}\frac{\Gamma(n_t+\epsilon_\gamma /T)}{\Gamma(\epsilon_\gamma /T)}\Big/ \frac{\Gamma(\epsilon_\gamma )}{\Gamma(N+\epsilon_\gamma-1)}\prod_{t=1}^{T}\frac{\Gamma(n_{t}^{-i}+\epsilon_\gamma /T)}{\Gamma(\epsilon_\gamma /T)}$}\\
&=\frac{\Gamma(N+\epsilon_\gamma -1)}{\Gamma(N+\epsilon_\gamma )} \frac{\Gamma(n_{t}+\epsilon_\gamma /T)}{\Gamma(n_{t}^{-i}+\epsilon_\gamma /T)}\\
&=\frac{n_{t}^{-i}+\epsilon_\gamma /T}{N+\epsilon_\gamma-1},
\end{aligned}
\end{equation}

and $p(y_{i}|Y^{-i},z_i = t,\bm{z}^{-i},\bm{c},\eta,\bm{\beta})$ is obtained as 
\begin{equation} \tag{7}
\begin{aligned}
&p(y_{i}|Y^{-i},z_i = t,\bm{z}^{-i},\bm{c},\eta,\bm{\beta})\\
&=\scalebox{0.8}{$
\prod_{c}\prod_{l\in c}\left [ \frac{\Gamma (\eta _{t}^{c})}{\Gamma (n_{lt}+\eta _{t}^{c})}\prod_{s}\frac{\Gamma (n_{lts}+\eta _{t}^{c}\bm{\beta} _{ts}^{c})}{\Gamma (\eta _{t}^{c}\bm{\beta}_{ts}^{c})}\Big /  \frac{\Gamma (\eta _{t}^{c})}{\Gamma (n_{lt}^{-i})+\eta _{t}^{c})}\prod_{s}\frac{\Gamma (n_{lts}^{-i})+\eta _{t}^{c}\bm{\beta} _{ts}^{c})}{\Gamma (\eta _{t}^{c}\bm{\beta}_{ts}^{c})}   \right ]$} \\
&=\prod_{c}\prod_{l\in c}\frac{\prod_{s}(n_{lts}^{-i}+\eta _{t}^{c}\bm{\beta} _{ts}^{c})^{I(y_{il}=s)}}{n_{lt}^{-i}+\eta _{t}^{c}}.
\end{aligned}
\end{equation}


\section{B}


a. Derivation of $p(\bm{\beta}_{tj}^c|\bm{\beta}_{t(\sim j)}^c,\eta _t^c,Y^c)$ is as follows:
\begin{equation} \tag{8}
\scalebox{0.9}{$
p(\bm{\beta}_{tj}^c|\bm{\beta}_{t(\sim j)}^c,\eta _t^c,Y^c)\propto p(Y^{cj}|Y^{-cj},\bm{\beta}_{t}^c,\eta _t^c)\times p(\bm{\beta}_{tj}^c|\bm{\beta}_{t(\sim j)}^c)$}.
\end{equation}

First, $\bm{\beta}_t^{c}$ follows dirichlet distribution:
\begin{equation} \tag{9}
\bm{\beta}_t^{c}|\bm{\alpha}_t\sim \mathrm{Dirichlet}({\bm{\alpha}}_t), 0< \bm{\beta}_{tj}^{c} <1, \sum_{j=1}^{T}\bm{\beta}_{tj}^{c} = 1,
\end{equation}
where the detailed form is given as:
\begin{equation}\tag{10}
\scalebox{0.9}{$
\left ( \bm{\beta}_{t1}^c ,...,\bm{\beta}_{tj}^c,...,\bm{\beta}_{tT}^c \right )\sim \mathrm{Dirichlet}\left ( \lambda_t \bm{\alpha} _{t1},...,\lambda_t \bm{\alpha} _{tj},...,\lambda_t \bm{\alpha} _{tT} \right )$}.
\end{equation}

With the aggregation property we obtain that 
\begin{equation} \tag{11}
\scalebox{0.9}{$
\left ( \bm{\beta}_{tj}^c ,u_{tj}^c,\bm{\beta}_{tt}^c=1- \bm{\beta}_{tj}^c- u_{tj}^c,\right )\sim \mathrm{Dirichlet}\left ( \lambda_t\bm{\alpha} _{tj},a_j,\lambda_t \bm{\alpha} _{tt} \right )$},
\end{equation}
where 
\begin{equation} \tag{12}
u_{tj}^c=1-\sum_{s=1,s\neq t,s\neq j} ^{T}\bm{\beta}_{ts}^{c},
\end{equation}
\begin{equation}\tag{13}
a_j=\sum_{s=1,s\neq t,s\neq j} ^{T}\lambda _t\bm{\alpha}_{ts}.
\end{equation}

The joint probability $p(\bm{\beta} _{tj}^c,u_{tj}^c)$ is given as
\begin{equation} \tag{14}
\begin{aligned}
&p(\bm{\beta} _{tj}^c,u_{tj}^c)=\prod_{l\in c}\frac{\Gamma(\sum_{j=1}^{T}\lambda_t\bm{\alpha}_{tj})}{\Gamma(\lambda_t\bm{\alpha}_{tj})\Gamma(a_j)\Gamma(\lambda_t\bm{\alpha}_{tt}) }\\
&\times(\bm{\beta} _{tj}^c)^{\lambda_t\bm{\alpha}_{tj}-1}(u _{tj}^c)^{a_j-1}(1-\bm{\beta} _{tj}^c-u _{tj}^c)^{\lambda_t\bm{\alpha}_{tt}-1}.
\end{aligned}
\end{equation}

The marginal probability $p(u_{tj}^c|\eta _t^c,Y^c)$ is computed as
\begin{equation} \tag{15}
\begin{aligned}
p(u_{tj}^c)=&\frac{\Gamma(\sum_{j=1}^{T}\lambda_t\bm{\alpha}_{tj})}{\Gamma(a_j)\Gamma(\lambda_t\bm{\alpha}_{tt}+\lambda_t\bm{\alpha}_{tj}) }\\
&\times(u _{tj}^c)^{a_j-1}(1-u _{tj}^c)^{\lambda_t\bm{\alpha}_{tt}+\lambda_t\bm{\alpha}_{tj}-1}.
\end{aligned}
\end{equation}

Then the conditional distribution $p(\bm{\beta}_{tj}^c|\bm{\beta}_{t(\sim j)}^c)$ is given as
\begin{equation} \tag{16}
p(\bm{\beta}_{tj}^c|\bm{\beta}_{t(\sim j)}^c)\propto \left ( \frac{\bm{\beta}_{tj}^c}{1-u_{tj}^c} \right )^{\lambda_t \bm{\alpha} _{tj}-1}\left ( 1- \frac{\bm{\beta}_{tj}^c}{1-u_{tj}^c}\right )^{\lambda_t \bm{\alpha}_{tt}-1}.
\end{equation}

According to Equation (3) in Appendix A, it can be easily obtained that 
\begin{equation} \tag{17}
\scalebox{0.9}{$
p(Y^{cj}|Y^{-cj},\bm{\beta}_{t}^c,\eta _t^c)=\prod_{l\in c}\left \{ \frac{\Gamma (n_{ltj}+\eta _t^c\bm{\beta}_{tj}^c)}{\Gamma (\eta _t^c\bm{\beta}_{tj}^c)} \frac{\Gamma (n_{ltt}+\eta _t^c\bm{\beta}_{tt}^c)}{\Gamma (\eta _t^c\bm{\beta}_{tt}^c)}\right \}$}.
\end{equation}

Finally we obtain that 
\begin{equation} \tag{18}
\begin{aligned}
&p(\bm{\beta}_{tj}^c|\bm{\beta}_{t(\sim j)}^c,\eta _t^c,Y^c)\propto p(Y^{cj}|Y^{-cj},\bm{\beta}_{t}^c,\eta _t^c)\times p(\bm{\beta}_{tj}^c|\bm{\beta}_{t(\sim j)}^c)\\
&\propto \prod_{l\in c}\left \{ \frac{\Gamma (n_{ltj}+\eta _t^c\bm{\beta}_{tj}^c)}{\Gamma (\eta _t^c\bm{\beta}_{tj}^c)} \frac{\Gamma (n_{ltt}+\eta _t^c\bm{\beta}_{tt}^c)}{\Gamma (\eta _t^c\bm{\beta}_{tt}^c)}\right \}\\&\times \left ( \frac{\bm{\beta}_{tj}^c}{1-u_{tj}^c} \right )^{\lambda_t \bm{\alpha}_{tj}-1}\left ( 1- \frac{\bm{\beta}_{tj}^c}{1-u_{tj}^c}\right )^{\lambda_t \bm{\alpha}_{tt}-1}.
\end{aligned}
\end{equation}

b. Derivation of $p(\eta _t^c|\bm{\beta}_t^c ,Y^c)$ is as follows:

The joint posterior distribution $p(\eta _t^c,\bm{\beta}_t^c |Y^c)$ is given as
\begin{equation} \tag{19}
\begin{aligned}
p(\eta _t^c,\bm{\beta}_t^c |Y^c)\propto \prod_{l\in c} &\frac{\Gamma(\eta _t^c)}{\Gamma (n_{lt}+\eta _t^c)}\prod_{j}\frac{\Gamma(n_{ltj}+\eta _t^c\bm{\beta }_{tj}^c)}{\Gamma (\eta _t^c\bm{\beta}_{t}^c)}\\
&\times \prod_{j}(\bm{\beta}_{tj}^c)^{\lambda_t\bm{\alpha}_{tj}-1}\times \lambda e^{-(\lambda\eta _t^c)}.
\end{aligned}
\end{equation}

Then the conditional posterior distribution $p(\eta _t^c|\bm{\beta}_t^c ,Y^c)$ is obtained with 
\begin{equation} \tag{20}
\scalebox{0.95}{$
p(\eta _t^c|\bm{\beta}_t^c ,Y^c)\propto \prod_{l\in c}\frac{\Gamma(\eta _t^c)}{\Gamma (n_{lt}+\eta _t^c)}\prod_{j}\frac{\Gamma(n_{ltj}+\eta _t^c\bm{\beta} _{tj}^c)}{\Gamma (\eta _t^c\bm{\beta} _{t}^c)}\times \lambda e^{-(\lambda\eta _t^c)}$}.
\end{equation}

\section{C. Metropolis-Hastings algorithm simulating $\bm{\beta}_{tj}^c$ and $\eta _t^c$}

\begin{algorithm}
    \caption{Metropolis-Hastings algorithm simulating $\bm{\beta}_{tj}^c$.}
    \label{alg:1}
    \begin{algorithmic}[1]

        \label{alg:1}
        \STATE Initialize $\bm{\beta}_{tj}^{c_0}$
        \FOR {$i=1,2,...$}
              \STATE Propose: $\bm{\beta}_{tj}^{c_\mathrm{cand}}\sim U(0,\min\left \{ 2\bm{\beta}_{tj}^{c_{i-1}} ,1\right \})$
              \STATE Acceptance probability: \\
              $\alpha (\bm{\beta}_{tj}^{c_\mathrm{cand}}|\bm{\beta}_{tj}^{c_{i-1}} )=\min\left \{ 1,\frac{p(\bm{\beta}_{tj}^{c_\mathrm{cand}}|\bm{\beta}_{t(\sim j)}^c,\eta _t^{c},Y^c)}{p(\bm{\beta}_{tj}^{c_{i-1}}|\bm{\beta}_{t(\sim j)}^c,\eta _t^c,Y^c)} \right \}$
               \STATE $ u\sim U(0,1)$
               \IF {$u<\alpha$}
               \STATE Accept the proposal: $\bm{\beta}_{tj}^{c_i} \leftarrow \bm{\beta}_{tj}^{c_\mathrm{cand}}$
               
               \ELSE
               \STATE Reject the proposal:$ \bm{\beta}_{tj}^{c_i} \leftarrow \bm{\beta}_{tj}^{c_{i-1}}$
               \ENDIF

         \ENDFOR

    \end{algorithmic}
\end{algorithm}

\begin{algorithm}
    \caption{Metropolis-Hastings algorithm simulating $\eta _t^c$.}
    \label{alg:2}
    \begin{algorithmic}[2]
 
        \label{alg:2}
        \STATE Initialize $\eta _t^{c_0}$
        \FOR {$i=1,2,...$}
              \STATE Propose: $\eta _t^{c_\mathrm{cand}}\sim U(0,\min\left \{ 2\eta _t^{c_{i-1}} ,1\right \})$
              \STATE Acceptance probability: \\
              $\alpha (\eta _t^{\mathrm{cand}}|\eta _t^{i-1})=\min\left \{ 1,\frac{p(\eta _t^{\mathrm{cand}}|\bm{\beta}_t^c ,Y^c)}{p(\eta _t^{i-1}|\bm{\beta}_t^c ,Y^c)} \right \}$
               \STATE $ u\sim U(0,1)$
               \IF {$u<\alpha$}
               \STATE Accept the proposal: $\eta _t^{c_i} \leftarrow \eta _t^{c_\mathrm{cand}}$
               
               \ELSE
               \STATE Reject the proposal:$ \eta _t^{c_i} \leftarrow \eta _t^{c_{i-1}}$
               \ENDIF
         \ENDFOR
    \end{algorithmic}
\end{algorithm}

\bibliography{ref}

\end{document}